\documentclass{article}

\usepackage{PRIMEarxiv}

\errorcontextlines\maxdimen

\usepackage[utf8]{inputenc} 
\usepackage[T1]{fontenc}    
\usepackage{url}            
\usepackage{booktabs}       
\usepackage{nicefrac}       
\usepackage{microtype}      
\usepackage{fancyhdr}       
\usepackage{graphicx}
\usepackage{xcolor}
\usepackage{threeparttable}
\usepackage{booktabs}
\usepackage{multirow}
\usepackage{tabularx}
\usepackage{arydshln}

\usepackage{lineno,hyperref}
\hypersetup{colorlinks, citecolor=blue, linkcolor=red, urlcolor=green}
\usepackage{amsmath,amsfonts,amssymb}
\usepackage{graphics}
\usepackage{bm}
\usepackage{multirow}
\usepackage{algorithm}
\usepackage{algpseudocode}

\usepackage{mathtools}
\usepackage{siunitx}
\DeclarePairedDelimiterXPP\BigOSI[2]%
  {\mathcal{O}}{(}{)}{}%
  {\SI{#1}{#2}}

\title{Towards Gaussian Process for operator learning: an uncertainty aware resolution independent operator learning algorithm for computational mechanics}

\author{Sawan Kumar \\
  Department of Applied Mechanics\\
  Indian Institute of Technology Delhi\\
  \texttt{sawan.kumar@am.iitd.ac.in} \\
  \And
  Rajdip Nayek\\
  Department of Applied Mechanics\\
  Indian Institute of Technology Delhi\\
  \texttt{rajdipn@am.iitd.ac.in} \\
  \And
  Souvik Chakraborty \\
  Department of Applied Mechanics\\
  Yardi School of Artificial Intelligence (ScAI)\\
  Indian Institute of Technology Delhi\\
  \texttt{souvik@am.iitd.ac.in} \\
}

\begin{document}
\maketitle

\begin{abstract}
The growing demand for accurate, efficient, and scalable solutions in computational mechanics highlights the need for advanced operator learning algorithms that can efficiently handle large datasets while providing reliable uncertainty quantification. This paper introduces a novel Gaussian Process (GP) based neural operator for solving parametric differential equations. The approach proposed leverages the expressive capability of deterministic neural operators and the uncertainty awareness of conventional GP. In particular, we propose a ``neural operator-embedded kernel'' wherein the GP kernel is formulated in the latent space learned using a neural operator. 
Further, we exploit a stochastic dual descent (SDD) algorithm for simultaneously training
the neural operator parameters and the GP hyperparameters. Our approach addresses the (a) resolution dependence and (b) cubic complexity of traditional GP models, allowing for input-resolution independence and scalability in high-dimensional and non-linear parametric systems, such as those encountered in computational mechanics. We apply our method to a range of non-linear parametric partial differential equations (PDEs) and demonstrate its superiority in both computational efficiency and accuracy compared to standard GP models and wavelet neural operators. Our experimental results highlight the efficacy of this framework in solving complex PDEs while maintaining robustness in uncertainty estimation, positioning it as a scalable and reliable operator-learning algorithm for computational mechanics.

\end{abstract}


\section{Introduction}
The solution of partial differential equations (PDEs) plays a crucial role in a wide range of scientific and engineering fields, including weather modeling \cite{pathak2022fourcastnet, lin2023spherical}, fracture mechanics \cite{Goswami_2022}, and medical diagnostics \cite{tripura2023wavelet_elastography}. For decades, classical solvers have been the primary method for solving PDEs, typically relying on mesh-based approaches to obtain solutions under specific initial and boundary conditions. However, these solvers often need to be re-executed for different conditions, leading to significant computational overhead.
In contrast, Neural Operators (NOs) \cite{li2023transformerpartialdifferentialequations,tripura2023wavelet,lu2021learning,li2020fourier} have emerged as a promising alternative to traditional solvers that allow learning the mapping between different function spaces. Neural operators are discretization invariant, enabling them to accommodate complex geometries and high-dimensional inputs while offering zero-shot super-resolution capabilities. Once trained, neural operators can efficiently predict solutions for varying initial and boundary conditions without the need for repeated runs, significantly reducing computational time compared to classical solvers.

The Deep Operator Network (DeepONet) is one of the pioneering neural operator architectures, introduced to address the challenge of learning operators between infinite-dimensional function spaces. Unlike conventional neural networks that map inputs to outputs in finite-dimensional spaces, DeepONet is a specially designed neural network architecture to model the relationship between entire functions, making it especially powerful for solving complex PDEs and related tasks. 
The core idea behind DeepONet is rooted in operator theory, where an operator maps one function to another. DeepONet consists of two main components: the ``branch network'' and the ``trunk network''. The outputs of both the branch and trunk networks are combined through an inner product resulting in the final output, which is a function value corresponding to the specific input. Variants of DeepONet include PCA-DeepONet \cite{LU2022114778}, physics-informed DeepONet \cite{goswami2022physicsinformeddeepneuraloperator}, and Nonlinear Manifold-based Operator (NOMAD) \cite{seidman2022nomadnonlinearmanifolddecoders} to name a few. However, most work on DeepONet is deterministic in nature and cannot capture the predictive uncertainty due to limited and noisy data. Development of probabilistic DeepONet is an active research area with recent attempts such as variational Bayes DeepONet (VB-DeepONet) \cite{garg2022variational}, $\alpha-$divergrence based DeepONet \cite{lone2024alphavideeponetpriorrobustvariational}, and conformal DeepONet \cite{moya2024conformalizeddeeponetdistributionfreeframeworkuncertainty}.

Parallel to the DeepONet-based architectures discussed above, kernel integration-based operator learning architectures have also gained popularity in recent times. These architectures are motivated by Green's function and utilize kernel integration to learn the mapping. Popular kernel integration-based neural operator includes Graph Neural Operator (GNO) \cite{li2020neuraloperatorgraphkernel}, Fourier Neural Operator (FNO) \cite{li2020fourier}, Laplace Neural Operator (LNO) \cite{cao2023lnolaplaceneuraloperator}, Wavelet Neural Operator (WNO) \cite{tripura2023wavelet},  and Convolutional Neural Operator (CNO) \cite{raonić2023convolutionalneuraloperatorsrobust} to name a few. Physics-informed variants of the operators can also be found in the literature \cite{N2024116546,li2023physicsinformedneuraloperatorlearning}. Kernel-based neural operators, being motivated by Green's function, are highly effective in capturing intricate patterns in data, making them efficient for a diverse array of applications \cite{rani2024generative,tripura2023wavelet_elastography,lin2023spherical,thakur2024mdnomadmixturedensitynonlinear}. However, kernel-based neural operators, by design are deterministic and cannot capture the predictive uncertainty. This constraint is crucial in applications where knowing the uncertainty of predictions is as essential as the predictions themselves; this is particularly true in domains such as medicine and atmospheric science where the availability of data is a challenge. 

Gaussian processes (GPs) have been around for quite a long time and provide a competitive probabilistic alternative to deep neural networks. GPs are non-parametric machine learning algorithm which has inherent capabilities to provide quantification of uncertainty. 
Researchers have also developed deep kernel learning \cite{wilson2015deep,ober2021promisespitfallsdeepkernel} for better expressiveness and physics-informed GP \cite{pförtner2023physicsinformed,chen2023sparse,besginow2022constraining} for training the model directly from known physics.
Although GPs are probabilistic models that are recognized for their robustness and capacity to give a measure of uncertainty, it has their own set of challenges. In particular, the learning capability of GPs is often constrained due to the limited choices of kernels available. Also, the training process for GPs involves computational complexity that grows cubically, making it computationally prohibitive to handle large datasets. Attempts such as the development of sparse GP \cite{snelson2005sparse,hensman2015scalable,chen2023sparse,uhrenholt2021probabilistic} can be found in the literature. However, scalability in GP is an open problem that often limits its application in practice.

In this work, we propose a novel probabilistic operator learning algorithm for solving computational mechanics problems. The proposed operator learning algorithm fuses the best of both neural operators and GP. We hypothesize that learning the kernel in latent space will allow better expressiveness. In particular, we propose to project the inputs to latent space by using neural operators and formulate the covariance kernel by using the latent variables. Note that with neural operators, it is possible to control the intrinsic dimensionality of the latent space and hence, the proposed model is discretization invariant. Although the proposed framework can be used with any neural operator, we here utilize WNO because of its proven performance. Further, to accelerate the training process, we propose to use a stochastic dual descent (SDD) algorithm for simultaneously training the neural operator parameters and the GP hyperparameters. We also propose a novel approach to initialize the SDD algorithm for better convergence. This facilitates the efficient handling of large data sets and provides reliable probabilistic solutions for parametric PDE. The key features of the proposed framework are summarized as follows:
\begin{itemize}
\item \textbf{Scalability and accuracy:} The proposed framework formulates the GP kernel in a latent space. This provides better expressiveness as is reflected in the results presented. Further, SDD-based algorithms proposed for simultaneously training the neural operator parameters and GP hyperparameters enhance the scalability of the model. 
\item \textbf{Uncertainty quantification:} The proposed framework naturally encompasses predictive uncertainty, making it well-suited for applications needing trustworthy uncertainty estimates.
\item \textbf{Discretization invariance:} The proposed framework is discretization invariant and can handle inputs of any discretization. This is achieved by exploiting the in-built discretization invariance feature of neural operators. 
\end{itemize}

The subsequent sections of this paper are organized as follows. The problem statement is discussed in Section \ref{PS}. Section \ref{Framework} explains the mathematical details of the proposed framework which includes the proposed architecture and the SDD optimization. Section \ref{Numercial_eg} demonstrates the implementation of our approach on a number of benchmark examples from the operator learning literature. Finally, Section \ref{Conclusion} provides a concluding review of the results and prospective future work in the study. 

\section{Problem statement}\label{PS}
Consider domain $\Omega \subseteq \mathbb{R}^{d_1}$, where $d_1$ represents the spatial dimension and can take values 1, 2, or 3 depending on whether it is a 1D, 2D, or 3D problem and $\mathcal{T}$ as the time period $\left(0, t_F\right)$. Furthermore, let $\bm{u}: \Omega \times \mathcal{T} \mapsto \mathbb{R}^{d_u}$ denote the solution to the specified initial boundary value problem (IBVP), 
\begin{equation}\label{eq:ge}
\begin{aligned}
\mathcal{L}(\bm{u};\bm{a})(\bm{x}, t) & =\bm{\varphi}(\bm{x}, t), & & \forall(\bm{x}, t) \in \Omega \times \mathcal{T} \\
\bm{u}(\bm{x}, 0) & =\bm{u}_0(\bm{x}), & & \forall \bm{x} \in \Omega \\
\bm{u}(\bm{x}, t) & =\bm{u}_{b c}(\bm{x}, t), & & \forall(\bm{x}, t) \in \partial \Omega \times \mathcal{T}
\end{aligned}
\end{equation}
where 
$\mathcal{L}$ is the differential operator, $\bm{\varphi}$ is the forcing term, and $\bm{u}_0$ is the initial condition. Let $\bm{z}: \Omega \times \mathcal{T} \mapsto \mathbb{R}^{d_z}$ represent the input to the system. The initial condition $\bm u_0$, boundary condition $\bm u_{bc}$, forcing function $\bm\varphi$, and parameter field $\bm a$ in Eq. \eqref{eq:ge} may be included in $\bm{z}$. 

If $\mathcal{Z}$ and $\mathcal{U}$ are Banach function spaces for input $\bm{z}$ and solution $\bm{u}$, then the integral operator $\overline{\mathcal{D}}: \mathcal{Z} \mapsto \mathcal{U}$ solves the IBVP using the solution operator $\overline{\mathcal{D}}$, 
\begin{equation}
    \bm{u} := \overline{\mathcal{D}}(\bm{z}).
\end{equation} 
In general, computing $\overline{\mathcal{D}}$ is computationally costly. Here, we seek solution via a neural operator $\mathcal{D} : \mathcal{Z} \times \bm{\Theta} \rightarrow \mathcal{U}$, given input $\bm{z} $ and parameters $\bm{\theta} \in \bm \Theta$. 
Given 
$N$-point training dataset, $\{\bm{z}_j, \bm{u}_j\}_{j=1}^N$, the unknown parameters $\bm \theta$ are generally obtained by minimising a error function $\mathcal E\left(\bm \theta\right)$,
\begin{equation}
    \bm \theta^* = \arg \min_{\bm \theta} \mathcal E\left(\bm \theta\right). 
\end{equation}
The above setup yields a deterministic neural operator that fails to account for the uncertainty due to limited and noisy training data. The objective of this study is to address this challenge by developing a probabilistic neural operator that  (a) is scalable, (b) yields accurate estimates, and (c) can quantify the predictive uncertainty because of limited and noisy training data.



\section{Proposed methodology}\label{Framework}
This section explains the foundation of the proposed framework, with all its essential components and mathematical formulations in detail. The key components of the proposed framework comprise the embedding WNO with vanilla kernels, stochastic training of the proposed framework using stochastic dual descent (SDD) optimization, and sampling from the Gaussian process' predictive distribution via pathwise sampling.
\subsection{Gaussian Process Operator}\label{framework}
We propose a Gaussian process-based neural operator, referred to as Gaussian Process Operator (GPO) which inherently quantifies uncertainty by learning a probabilistic mapping between input-output functions.
Considering the training dataset defined in Section \ref{PS}, the input-output mapping is represented as

\begin{equation}
\bm u = \bm{f}_{\theta}(\bm{z}) + \bm{\epsilon}, \quad \bm{\epsilon} \sim \mathcal{N}(\bm{0}, \sigma_n^2 \mathbf{I})
\end{equation}
where $\bm{\epsilon}$ represents the noise term with $\sigma_n^2$ as the noise variance and $\bm z$ consists of the discretized parametric field along with the grid information $\bm x \in \Omega_d$. 
Similar to conventional GP, we assign a GP prior to $\bm{f}(\bm z)$,
\begin{equation}\label{eq:gpprior}
\bm{f}(\bm {z}) \sim \mathcal{GP}(\bm{0}, \mathbf{K}(\bm{z}, \bm{z}'))
\end{equation}
where $\mathbf{K}(\bm{z}, \bm{z}')$ is the covariance kernel. In this work, \textit{we propose to formulate the covariance kernel in the latent space by projecting the inputs into the latent space using a neural operator}. The key idea is to replace the vanilla kernel $k(\cdot,\cdot)$ (e.g. family of Mat\'{e}rn kernels, RBF kernel) with a neural operator embedded kernel utilizing an existing vanilla kernel as the base kernel. Although the proposed framework can be used with any neural operator, we have considered WNO as the choice of neural operator due to its better performance across a wide variety of PDEs. Details on WNO are provided in \ref{appendix_WNO}. The resulting kernel designed with WNO can be constructed as follows:
\begin{equation}\label{eq:wno_kernel}
    k^{\text{wno}}(\bm{z}_i, \bm{z}_j; \bm \theta) 
    = k(\psi_{\bm\theta_1}(\bm{z}_i), \psi_{\bm\theta_1}(\bm{z}_j); \bm \theta_2), \quad  \forall i,j = 1,\ldots,N,
\end{equation}
where $\bm{z}_i,\bm{z}_j$ denotes any input to the covariance kernel , $\psi_{\bm\theta_1}(\cdot)$ denotes the neural operator parametrized by $\bm\theta_1$, and $\bm \theta_2$ represents the hyperparameters of the kernel $k$, including length-scale parameter $l$ and process variance $\sigma^2$. It is important to note that, the neural operator parameters and the GP hyperparameters are unknown apriori, and are to be obtained by minimizing a loss function. One example of WNO-embedded kernel constructed using  Mat\'{e}rn-5/2 is as follows:
\begin{equation}\label{eq:Matern Kernel eq}
\begin{split}
    k^{\text{wno}}(\bm{z}_i, \bm{z}_j; \bm \theta) = & k_{5 / 2, l}\left(\psi_{\bm \theta_1}(\bm{z}), \psi_{\bm \theta_1}(\bm{z}^{\prime}\right); \bm \theta_2)
    = \\ & \sigma_n^2 \left(1+\frac{\sqrt{5}\left\|\psi_{\bm \theta_1}(\bm{z})-\psi_{\bm \theta_1}(\bm{z}^{\prime})\right\|_2}{l}+\frac{5\left\|\psi_{\bm \theta_1}(\bm{z})-\psi_{\bm \theta_1}(\bm{z}^{\prime})\right\|_2^2}{3 l^2}\right) \exp \left(-\frac{\sqrt{5}\left\|\psi_{\bm \theta_1}(\bm{z})-\psi_{\bm \theta_1}(\bm{z}^{\prime})\right\|_2}{l}\right).
\end{split}
\end{equation}
Here, $\|\psi_{\bm \theta_1}(\bm{z}) - \psi_{\bm \theta_1}(\bm{z}^{\prime})\|_2$ represents the Euclidean distance in the latent space, and $\bm \theta_2 = \left[l, \sigma^2\right]$ represent the GP hyperparameter. The neural operator-embedded kernel construction allows the GP to inherit the neural operators' learning capability by improving the vanilla kernels' expressiveness, enabling it to learn complex patterns in the data.

\subsection{Training optimization using Stochastic dual descent}
Considering Eq. \eqref{eq:gpprior} with WNO-embedded kernel in Eq. \eqref{eq:wno_kernel}, we have, 
\begin{equation}
\bm f(\bm{z}) \sim \mathcal{G}\mathcal{P}(\bm{0}, \mathbf{K}^{wno}(\bm{z}, \bm{z}';\bm{\theta})).
\end{equation}
Given the training dataset, the parameters of the neural operator and the hyperparameters in GP kernel $\bm{\theta}$ can be trained by minimizing the negative log-likelihood; however, this approach has a time complexity that increases cubically with the number of training samples as for $N$ training samples the Gram matrix becomes $N\times N$ which needs to be inverted and inversion operation has cubic complexity. Additionally, the memory requirement for an $N\times N$ matrix scales with complexity $\mathcal{O}(N^2)$ meaning that as the matrix size increases, the memory usage grows quadratically. Using the optimized hyperparameters, the predictive distribution is computed as
\begin{subequations}
    \begin{equation}
        \bm{f} \mid \bm{U} \sim \mathcal{GP}(\bm\mu_{\bm{f} \mid \mathbf{U}}, \bm{K}^{wno}_{\bm{f} \mid \mathbf{U}})
    \end{equation}
    \begin{equation}
       \bm\mu_{\bm f \mid \mathbf{U}}(\cdot) = \bm{k}^{wno}(\cdot, \mathbf{Z};\bm{\theta})(\mathbf{K}^{WNO}_{\mathbf{ZZ}} + \sigma_n^2 \mathbf{I})^{-1} \mathbf{U}, 
    \end{equation}
\end{subequations}
where $\bm\mu_{\bm{f} \mid \mathbf{U}} \in \mathbb{R}^{d_u}$ is the posterior mean function, $\mathbf{K}^{wno}_{\bm{f} \mid \mathbf{U}}$ is the posterior covariance function, and $\mathbf{K}^{wno}_{\mathbf{ZZ}}$ is the kernel matrix $\left[k^{wno}(\bm{z}_i,\bm{z}_j;\bm{\theta})\right]_{i,j}$. One can refer to \cite{Rasmussen2004, bishop2006pattern} for a detailed explanation.

In response to the computational cost discussed above, we employ stochastic dual descent (SDD) for the training of the proposed model. Using SDD, we 
minimize an objective that is similar to that obtained in kernel ridge regression. The primal objective function \cite{lin2024samplinggaussianprocessposteriors} denoted by $L_{p}(\mathbf{A})$ is defined as:
\begin{equation}\label{primal_form}
L_{p}(\mathbf{A}) = \frac{1}{2} \|\mathbf{U} - \mathbf{K} \mathbf{A}\|_2^2 + \frac{\sigma_n}{2} \|\mathbf{A}\|^2_{\mathbf{K}}
\end{equation}
\begin{equation} 
\nabla_{\mathbf{A}}L_p(\mathbf{A})=\mathbf{K}(\sigma_n\mathbf{A} - \mathbf{U} + \mathbf{K}\mathbf{A}) 
\end{equation} 
\begin{equation} 
\nabla^2_{\mathbf{A}} L_p(\mathbf{A}) = \mathbf{K}(\mathbf{K} + \sigma_n\mathbf{I})
 \end{equation}
where $\nabla_{\mathbf{A}} L_p(\mathbf{A})$, $\nabla^2_{\mathbf{A}} L_p(\mathbf{A})$ represents the gradient and Hessian of the primal objective function, \(\mathbf{A} \in \mathbb{R}^{N \times d_u}\) is the representer weight matrix which reflects the relative contribution of each training point's kernel function on the prediction for a new point, \(\mathbf{U} \in \mathbb{R}^{N \times d_u}\) is the matrix of observed targets, $\mathbf{K}$ is the kernel matrix with entries \(k^{wno}(\mathbf{z}_i, \mathbf{z}_j)\), and \(\sigma_n^2 > 0\) is the variance parameter. The term \(\|\mathbf{A}\|^2_{\mathbf{K}}\) represents the \(\mathbf{K}\)-weighted norm, defined as \(\|\mathbf{A}\|^2_{\mathbf{K}} = \text{tr}(\mathbf{A}^T \mathbf{K} \mathbf{A})\), where \(\text{tr}(\cdot)\) denotes the trace of the matrix. Typically, in the formulation of Bayesian linear models and GPs, the approach that considers the weights is commonly referred to as the primal form, whereas the perspective that involves kernels is known as the dual form. However, in the context of kernel methods, a different terminology is used. In kernel methods, approaches that work directly with entities residing in the reproducing kernel Hilbert space (RKHS) are typically called the dual form, as they avoid explicit feature mapping and instead operate using kernel functions. 

The primal form in Eq. \eqref{primal_form} often suffers from numerical instability during training, especially when larger step sizes are used leading to slow convergence. In order to overcome these challenges, a dual objective is introduced in \cite{lin2024stochasticgradientdescentgaussian} which allows for larger step sizes and faster, more stable convergence. The dual objective has the same unique minimizer as the primal objective. The dual objective \(L_{d}(\mathbf{A})\) is given by:
\begin{equation}\label{eq:sdd}
L_{d}(\mathbf{A}) = \frac{1}{2} \|\mathbf{A}\|^2_{\mathbf{K} + \sigma_n \mathbf{I}} - \text{tr}(\mathbf{A}^T \mathbf{U})
\end{equation}
where \(\mathbf{I} \in \mathbb{R}^{N \times N}\) is the identity matrix.
The dual gradient \(\mathbf{G}(\mathbf{A})\) and Hessian \(\nabla^2 L_{d}(\mathbf{A})\) are:
\begin{subequations}
\begin{equation}
\mathbf{G}(\mathbf{A}) = (\mathbf{K} + \sigma_n \mathbf{I}) \mathbf{A} - \mathbf{U}
\end{equation}
\begin{equation}
\nabla^2 L_{d}(\mathbf{A}) = \mathbf{K} + \sigma_n \mathbf{I}
\end{equation}
\end{subequations}
Employing the gradient of the dual objective $\nabla{L_{d}}$ allows us to use a bigger step size thus making the optimization more stable and achieving faster convergence.  
We exploit Eq. \eqref{eq:sdd} to train the proposed model. By using the SDD algorithm, we ensure that the computational complexity remains manageable, even for large datasets, due to the utilization of mini-batches and efficient gradient updates. The training algorithm for the optimization of the proposed framework is given in Algorithm \ref{alg_sdd}. 
\begin{algorithm}[ht!]
\caption{Training of the Proposed Framework}
\label{alg_sdd}
\begin{algorithmic}[1]
\Require Kernel matrix $\mathbf{K}$ with rows $\bm{k}_1, \ldots, \bm{k}_N$, targets $\mathbf{U} \in \mathbb{R}^{N \times d_u}$, likelihood variance $\sigma_n > 0$, number of steps $T \in \mathbb{N}^+$, batch size $B \in \{1, \ldots, N\}$, step size $\beta > 0$, momentum parameter $\rho \in [0, 1)$, averaging parameter $r \in (0, 1]$
\Ensure Approximation of $\bar{\mathbf{A}} = (\mathbf{K} + \sigma_n \mathbf{I})^{-1} \mathbf{U}$
\State $\mathbf{V}_0 = 0$; $\mathbf{A}_0 = 0$; $\bar{\mathbf{A}}_0 = 0$ \Comment{Initialize all in $\mathbb{R}^{N \times d_u}$}
\For{$t \in \{1, \ldots, T\}$}
\State Sample $\mathcal{I}_t = \{i_{t1}, \ldots, i_{tB}\} \sim \text{Uniform}(\{1, \ldots, N\})$ \Comment{Random batch indices}
\State Compute the WNO-embedded kernel:
\State $\bm{\psi}_{\bm{\theta_1}}(\bm{z}) \leftarrow \bm{\psi}_{\bm{\theta}_1}(\mathbf{Z}_{\mathcal{I}_t})$ \Comment{Use WNO transformation to inputs}
\State $\mathbf{K}^{\text{wno}} \leftarrow k(\bm{\psi}_{\bm{\theta_1}}(\bm{z}), \bm{\psi}_{\bm{\theta_1}}(\bm{z'}))$ \Comment{ComputeWNO-embedded kernel matrix}
\State Compute Gradient Estimate:
\State $\mathbf{G}_t = \frac{N}{B} \sum_{i \in \mathcal{I}_t} \left[ (\mathbf{K}^{\text{WNO}}_i + \sigma_n \mathbf{e}_i)^T (\mathbf{A}_{t-1} + \rho \mathbf{V}_{t-1}) - \mathbf{U}_i \right] \mathbf{e}_i$
\State Update Velocity:
\State $\mathbf{V}_t = \rho \mathbf{V}_{t-1} - \beta \mathbf{G}_t$
\State Update Parameters:
\State $\mathbf{A}_t = \mathbf{A}_{t-1} + \mathbf{V}_t$
\State Iterate Averaging:
\State $\bar{\mathbf{A}}_t = r \mathbf{A}_t + (1 - r) \bar{\mathbf{A}}_{t-1}$ \Comment{Geometric averaging}
\State Compute Loss:
\State $L_t(\mathbf{A}) = \frac{1}{2} \|\mathbf{U} - \mathbf{K} \mathbf{A}_t\|^2 + \frac{\sigma_n}{2} \|\mathbf{A}_t\|_{\mathbf{K}}^2$
\EndFor
\State \Return $\bar{\mathbf{A}}_T$
\end{algorithmic}
\end{algorithm}

\begin{figure}[ht!]
\centering
\includegraphics[width=\textwidth]{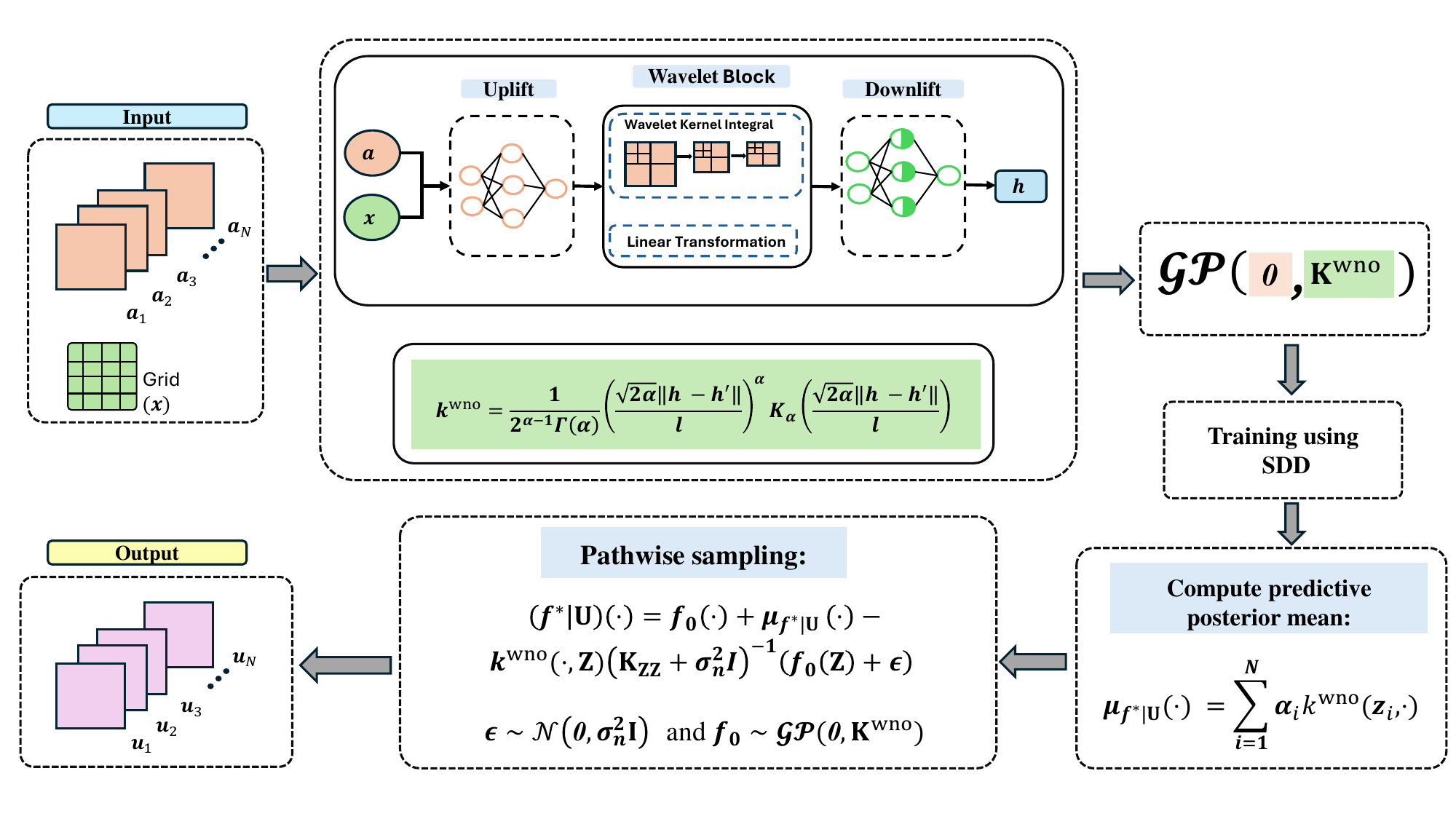}
\caption{\textbf{Schematics of Gaussian Process operator}. The architecture of the proposed framework is depicted in the above schematic diagram, which provides an overview of the framework. The schematic illustrates the transformation of inputs in the latent space through the utilization of a WNO-embedded kernel. Following this, stochastic dual descent (SDD) optimization is utilized for the training of the framework. The proposed approach transforms the training data via the WNO-embedded kernel thus improving the learning capabilities of the vanilla kernels, and utilizes SDD for efficient optimization. The posterior mean and pathwise sampling steps provide the final predictions with the associated predictive uncertainty.
}
\label{fig_framework}
\end{figure}

\subsection{GP posterior inference and predictive posterior sampling }
Given the optimal representer weights $\mathbf{A}^* = \{\bm{\alpha}_1,\ldots,\bm{\alpha}_N \}^{T} \in \mathbb{R}^{N \times d_u}$, we can compute the predictive posterior mean of the GP using the representer theorem \cite{bohn2019representer}. For a test input $\bm{z}_* $, the predictive posterior mean $\mu_{\bm{f}^* | \mathbf{U}}(\bm{z}_*)$ is computed as follows:
\begin{equation}
\bm\mu_{\bm{f}^* | \mathbf{U}}(\bm{z}_*) = \bm{k}^{\text{wno}}(\bm{z}_*, \mathbf{Z};\bm{\theta}) (\mathbf{K}_{\mathbf{ZZ}} + \sigma_n^2 \mathbf{I})^{-1} \mathbf{U} = \bm{k}^{\text{wno}}(\bm{z}_*, \mathbf{Z};\bm{\theta}) \mathbf{A}^*
\end{equation}
Here, $\bm{k}^{\text{wno}}(\bm{z}_*, \mathbf{Z};\bm{\theta}) \in \mathbb{R}^{1 \times N}$ is the WNO-embedded kernel matrix evaluated between the test inputs and training inputs, $\mathbf{A}^* \in \mathbb{R}^{N \times d_u}$ are the optimal representer weights, and $\mathbf{U} \in \mathbb{R}^{N \times d_u}$ is the matrix of observed targets.
Pathwise sampling \cite{wilson2021pathwiseconditioninggaussianprocesses} allows us to generate samples from the GP predictive posterior, and is particularly useful for scenarios where we want to understand the uncertainty over the function space. Using pathwise sample we get the following expression for the posterior:
\begin{equation}\label{pathwise_sampling}
(\bm{f}^* | \mathbf{U})(\cdot) = 
\underbrace{\bm{f}(\cdot)}_{\text{prior sample}} + 
\underbrace{\bm{\mu}_{\bm{f}^* | \mathbf{U}}(\cdot)}_{\text{predictive mean}} - 
\underbrace{\bm{k}^{\text{wno}}(\cdot, \mathbf{Z};\bm{\theta})(\mathbf{K}_{\mathbf{ZZ}} + \sigma_n^2\mathbf{I})^{-1} (\bm{f}(\mathbf{Z}) + \mathbf{\epsilon})}_{\text{uncertainty reduction term}}
\end{equation}
where $\bm{\epsilon} \sim \mathcal{N}(\bm{0}, \sigma_n^2\mathbf{I})$ and $\bm{f} \sim \mathcal{GP}(\bm{0}, \mathbf{K}^{\text{wno}})$. The formulation in Eq. \eqref{pathwise_sampling} leverages the structure of the GP and the properties of the kernel function \(k^{\text{wno}}(\bm{z}_i, \cdot)\) to efficiently compute both the mean and samples from the predictive posterior distribution. The first term of Eq. \eqref{pathwise_sampling} represents a sample from the GP prior before observing any data. The second term reflects the mean estimate of the function after conditioning on the training data and the uncertainty reduction term captures how much uncertainty is reduced by incorporating the training observations. It adjusts the prior function sample $\bm f(\cdot)$ based on the observed training data.
First, we sample from the prior $\bm{f} \sim \mathcal{GP}(\bm{0}, \mathbf{K}^{\text{wno}})$.
The predictive posterior mean using the optimal representer weights $\mathbf{A}^*$ is then computed as,
\begin{equation}
\bm\mu_{\bm{f}^* | \mathbf{U}}(\bm{z}_*) = \sum_{i=1}^{N} \bm{\alpha}_i k^{\text{wno}}(\bm{z}_i, \bm{z}_*)
\end{equation}
where $\bm{\alpha}_i \in \mathbb{R}^{d_u}$ are the rows of the representer weight matrix $\mathbf{A}^*$.
Next, we compute the uncertainty term using the SDD (Algorithm \ref{alg_sdd}) again to obtain another set of representer weights $\mathbf{B}$ where,
\begin{equation}
\mathbf{B} = (\mathbf{K} + \sigma_n \mathbf{I})^{-1} (\bm{f}(\mathbf{Z}) + \mathbf{\epsilon})
\end{equation}
The uncertainty term can be computed using optimal representer weights $\mathbf{B}^* = \{\bm{\beta}_1,\ldots,\bm{\beta}_N \}^{T} \in \mathbb{R}^{N \times d_u}$,
\begin{equation}
\bm{\mu}_{\mathbf{B}}(\bm{z}_*) = \sum_{i=1}^{N} \bm{\beta}_i k^{\text{wno}}(\bm{z}_i, \bm{z}_*)
\end{equation}
Using the conditional expression for the pathwise sampling represented in Eq. \eqref{pathwise_sampling}, we can compute the predictive posterior samples using the expression:
\begin{equation}
\bm{f}_{\text{predictive}}^*(\bm{z}_*) = 
\underbrace{\bm{f}(\bm{z}_*)}_{\text{prior sample}} + 
\underbrace{\mu_{\bm{f}^* | \mathbf{U}}(\bm{z}_*)}_{\text{predictive mean}} - 
\underbrace{\mu_{\mathbf{B}}(\bm{z}_*)}_{\text{uncertainty reduction term}}
\end{equation}
This method ensures accurate estimation of the posterior mean and uncertainty, resulting in reliable predictions with quantifiable uncertainty. Details on the inference algorithm are provided in Algorithm \ref{alg_inference}. 
The overall framework is schematically shown in Fig. \ref{fig_framework}.

\begin{algorithm}
\caption{Inference: Pathwise Sampling}
\label{alg_inference}
\begin{algorithmic}[1]
\Require Test input $\bm{z}_*$, Kernel function $k^{\text{wno}}$, targets $\mathbf{U} \in \mathbb{R}^{N \times d_u}$, prior sample $\bm{f} \sim \mathcal{GP}(\bm{0}, \mathbf{K}^{\text{wno}})$, noise $\mathbf{\epsilon} \sim \mathcal{N}(\bm{0}, \sigma_n \mathbf{I})$, learned represented weights $\mathbf{A}^*$
\Ensure Samples from the predictive posterior of GP
\State Compute the predictive posterior mean: $\bm{\mu}_{\bm{f}^* | \mathbf{U}}(\bm{z}_*) = \bm{k}^{\text{wno}}(\bm{z}_*, \mathbf{Z}) \mathbf{A}^*$
\State Compute the uncertainty term using new representer weights $\mathbf{B}$: $\bm{\mu}_{\mathbf{B}}(\bm{z}_*) = \sum_{i=1}^{N} \bm{\beta}_i k^{\text{wno}}(\bm{z}_i, \bm{z}_*)$
\State Compute the predictive posterior sample: $\bm{f}_{\text{predictive}}^*(\bm{z}_*) = \bm{f}(\bm{z}_*) + \bm{\mu}_{\bm{f}^* | \mathbf{U}}(\bm{z}_*) - \bm{\mu}_{\mathbf{B}}(\bm{z}_*)$
\State \Return $\bm{f}_{\text{predictive}}^*$
\end{algorithmic}
\end{algorithm}

\section{Numerical Examples}\label{Numercial_eg}

In this section, we illustrate a variety of case studies to test the performance of the proposed Gaussian Process Operator (GPO). The numerical examples range from various benchmark problems for neural operator learning frameworks like (a) 1D Burger's equation, (b) 1D wave-advection equation,(c) 2D Darcy flow equation, (d) 2D Darcy flow equation with a notch in the triangular domain, and (e) 2D Navier Stokes equation. A relative $L_2$ loss for the test set is used to estimate the performance of the proposed framework. In addition, a comparative study is done to evaluate the performance of the proposed framework with GPO (without SDD), vanilla WNO, and GP (with the vanilla kernel). The framework's performance is summarized in the Table \ref{table_accuracy}. In regard to the architecture of the WNO-embedded kernel design, it is composed of a Matern kernel as the base kernel while WNO has layers varying from 2 to 3 in accordance with the example in hand. The ADAM optimizer is employed for the training of the proposed framework. The other hyperparameter settings encompass the batch size, learning rate, geometric averaging term, and number of epochs. Table \ref{table_experimental_settings} provides a comprehensive summary of the experimental and hyperparameter settings used for the various numerical examples for the proposed framework. Further details of individual cases are discussed with respective subsections.

\begin{table}[ht!]
\centering
\caption{Experimental settings for numerical examples and case study}
\label{table_experimental_settings}
\begin{threeparttable}
\small 
\begin{tabular}{lcccccccc}
\hline
\textbf{Example} & \textbf{Training} & \textbf{Testing} & \textbf{Batch} & \textbf{Levels of wavelet} & \textbf{Learning} & \textbf{Averaging} & \textbf{Epochs} \\
\textbf{} & \textbf{samples} & \textbf{samples} & \textbf{size} & \textbf{decomposition} & \textbf{rate} & \textbf{term} & \textbf{} \\
\hline
Burgers$^{(\ref{sec:bur_eq})}$ & 1100 & 200 & 32 & 8 & $10^{-1}$ & 0.9 & 450\\
Wave advection$^{(\ref{sec:AD_eq})}$ & 1000 & 200 & 20 & 6 & $10^{-1}$ & 0.9 & 400\\
Darcy$^{(\ref{sec:Darcy_eq})}$  & 1000 & 200 & 16 & 4 & $10^{-2}$ & 0.8 & 300 \\
Darcy (with notch)$^{(\ref{sec:darcy_td})}$ & 1000 & 200 & 8 & 3 & $10^{-3}$ & 0.8 & 200 \\
Navier-Stokes$^{(\ref{sec:NS_eq})}$ & 3000 & 200 & 16 & 3 & $10^{-1}$ & 0.8 & 200 \\
\hline
\end{tabular}
\end{threeparttable}
\end{table}


\subsection{1D Burger Equation}\label{sec:bur_eq}
\begin{figure}[ht!]
	\centering
        \includegraphics[width=\textwidth]{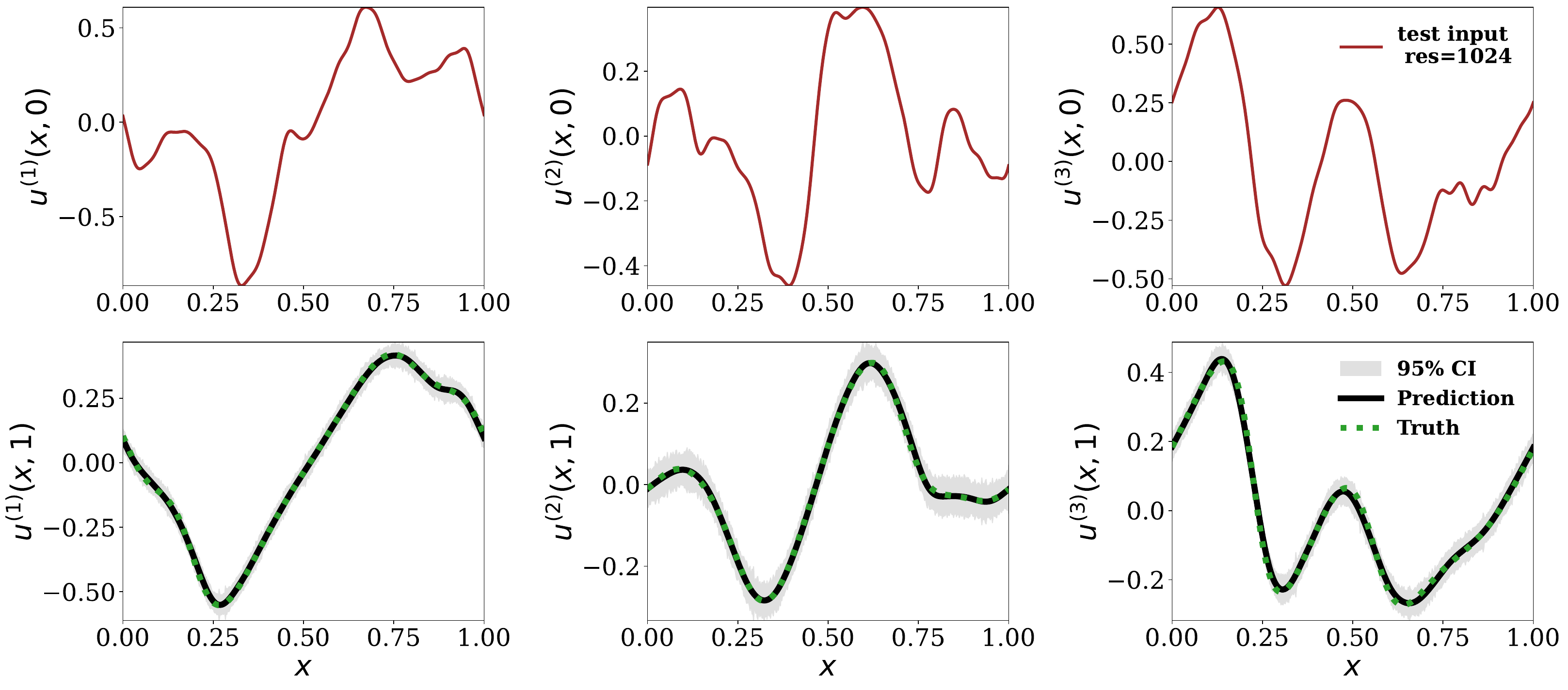}
	\caption{\textbf{1D Burger:} Figure shows the prediction obtained from the proposed framework for three of the representative test inputs. The first row shows one of the three representative test inputs (initial conditions) from the test dataset, while the second row shows the corresponding ground truth and the mean prediction obtained from our proposed framework at time $t=1$ along with a 95 \% confidence interval.}
	\label{fig_burger}
\end{figure}

\begin{figure}[ht!]
	\centering
        \includegraphics[width=\textwidth]{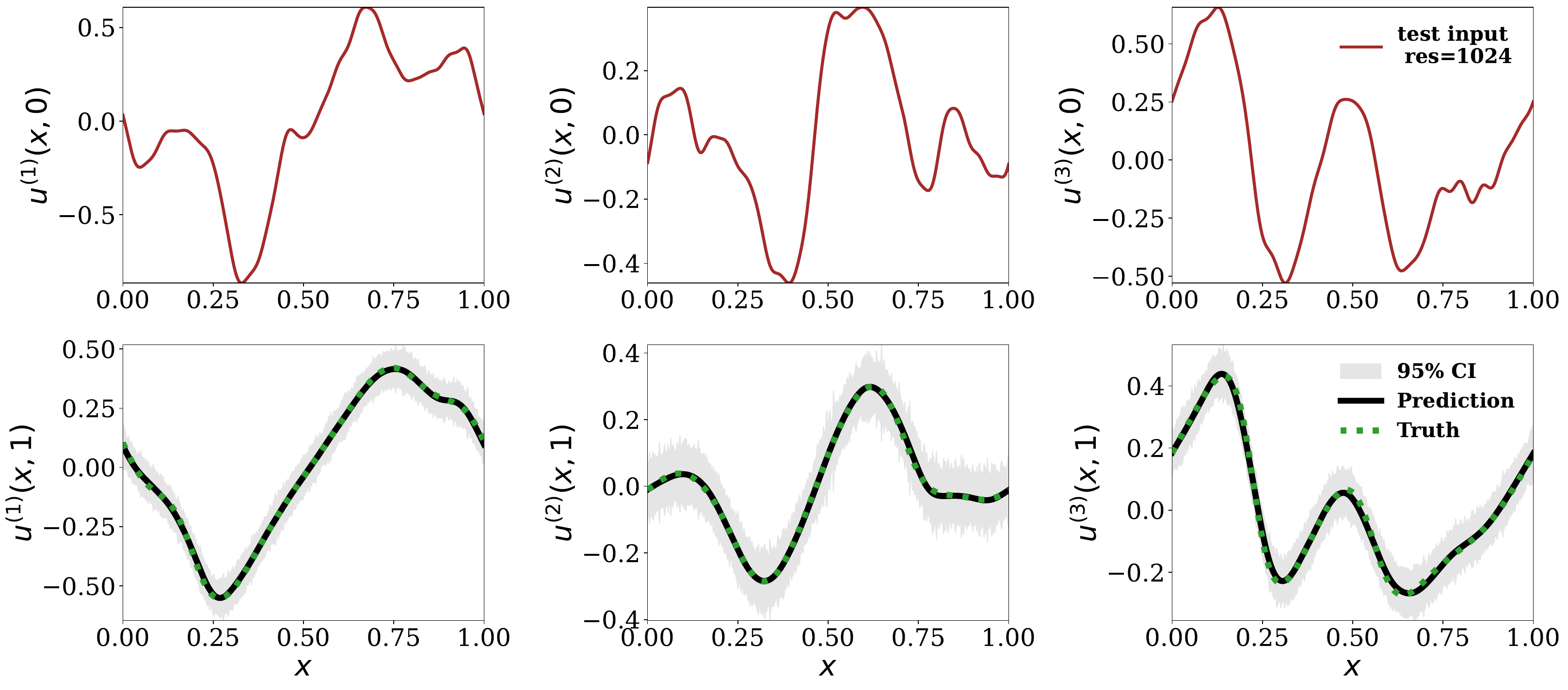}
	\caption{\textbf{1D Burger:}Figure shows the prediction obtained from the proposed framework for three of the representative test inputs. The first row shows one of the three representative test inputs on a spatial resolution of 1024 which is higher than the training input resolution of 512. The second row shows the corresponding ground truth and the mean prediction obtained from our proposed framework along with a 95 \% confidence interval.}
	\label{fig_burger_sup}
\end{figure}

The 1D Burgers equation is a widely utilized partial differential equation that represents one-dimensional flows in diverse domains, such as fluid mechanics, gas dynamics, and traffic flow. An interesting feature of this equation is that it exhibits both diffusive and advective behavior, making it useful for studying a wide range of phenomena. The mathematical expression of the 1D Burger equation with periodic boundary conditions is as follows:
\begin{equation}\label{eq_BS}
\begin{aligned}
\partial_{t} u(x, t) + 0.5\partial_{x} u^{2}(x,t) &= \nu \partial_{x x} u(x,t), & & x \in(0,1), t \in(0,1] \\
u(x = 0,t) &= u(x= 1,t), & & x \in(0,1), t \in(0,1] \\
u(x,0) &= u_{0}(x), & & x \in(0,1).
\end{aligned}
\end{equation}
Here, $\nu$ is the viscosity of the flow, which is positive and greater than zero, and $u_{0}(x) $ is the initial condition. The initial conditions $u_{0}(x)$, are generated using a Gaussian random field where $u_{0}(x) \sim \mathcal{N}(0,625(- \Delta + 25I)^{-2}).$ The aim is to learn the mapping from input to output function spaces for the 1D Burger's PDE, $ u(x, t = 0) \mapsto u(x, t = 1) $. We consider $\nu = 0.1$ and a spatial resolution of 512. The dataset utilized for training and testing is sourced from Ref. \cite{li2020fourier}. The proposed framework utilizes 1100 instances of training inputs followed by 200 instances of test inputs. The parameters of the GPO were initialized using the optimized parameters obtained from the exact inference of the proposed operator from a subset of the training samples. This ensures a reasonable initial condition that aid towards proper convergence using SDD. 
For SDD, we utilized a batch size of 32, a learning rate of 0.1, and performed a total of 450 epochs. Additional details regarding the experimental settings are provided in Table \ref{table_experimental_settings}.
Fig. \ref{fig_burger} illustrates the predictions obtained from the proposed framework. It displays predictions for three different test inputs, with the ground truth obtained from a numerical solver and the mean predictions from our proposed framework. Fig. \ref{fig_burger} indicates that our proposed framework can produce accurate predictions along with predictive uncertainty estimates. A relatively tight envelope of the predictive uncertainty indicates the model's confidence in its prediction. Additionally, the fact that the ground truth solution is always within the confidence interval illustrates that the learned model is not overconfident. 
Overall, we observe that the proposed framework outperforms the vanilla WNO and GP with vanilla kernel in terms of predictive accuracy (see Table \ref{table_accuracy}). Utilizing SDD allows the proposed GPO to scale with an increase in the number of training samples. This results in superior performance of GPO as opposed to GPO (without SDD).
In Fig. \ref{fig_burger_sup}, our proposed framework is also tested on inputs of higher spatial resolution compared to the training samples. We can observe that our proposed framework, without any training, can handle inputs of higher resolution illustrating the discretization invariance of the proposed framework. We note that the uncertainty bound increases for finer resolution input. This indicates that the model becomes less confident as we move away from the training resolution. This can be attributed to the fact that finer resolution corresponds to input at new grid locations.

\begin{figure}[ht!] 
	\centering 
        \includegraphics[scale=0.40]{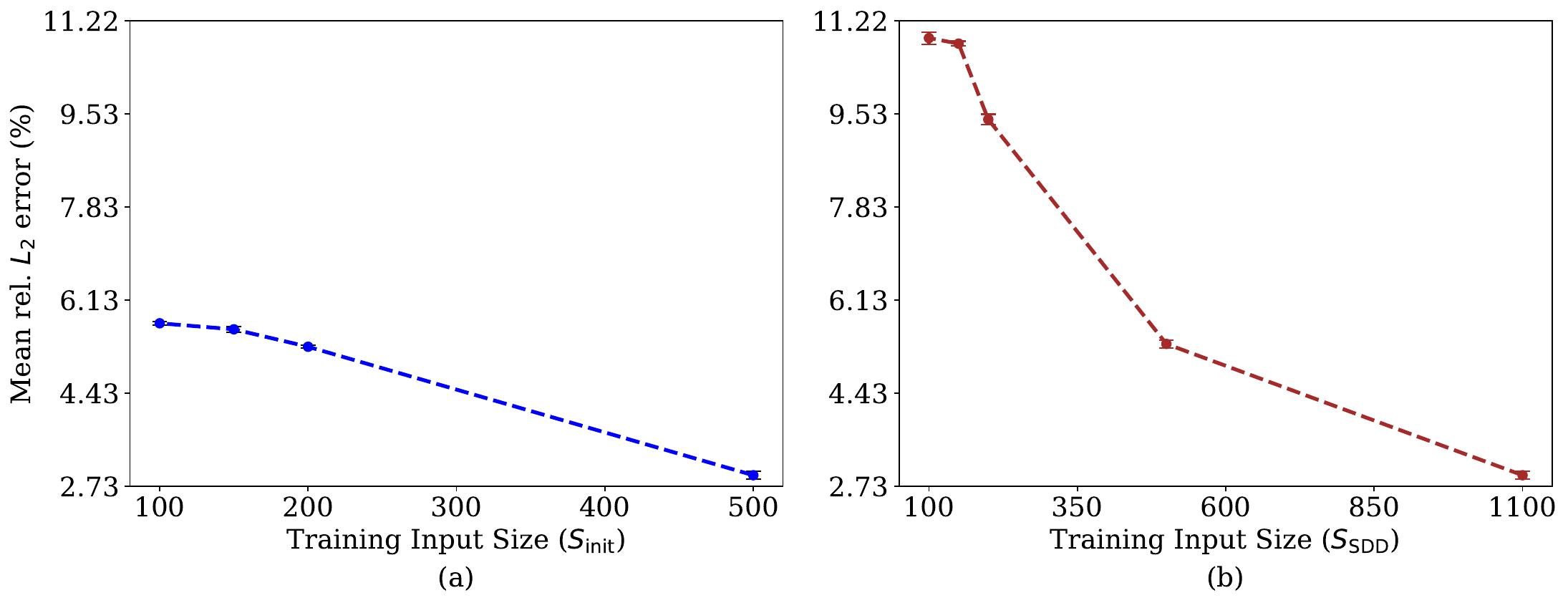} 
	\caption{\textbf{1D Burger}. Figure (a) shows the effect of initialization on the performance of the proposed model. We vary the number of samples $S_{\text{init}}$ used for initializing the model while keeping overall training samples $S_{\text{SDD}} = 1100$.  Figure (b) illustrates the effect of the number of training samples $S_{\text{SDD}}$ on the performance of the proposed approach. For all the cases, the model is initialized with $S_{\text{init}}=500$.} 
	\label{fig_cs_burger} 
\end{figure}

We further conducted case studies to understand the effect of hyperparameters on the performance of the proposed framework. In particular, we investigate the effect of initialization and the number of training samples on the performance of the proposed framework. As previously stated, the proposed model is initialized by first training it conventionally on a subset of samples $\left(S_{\text{init}} \right)$. Therefore, the effect of initialization is investigated by varying  $S_{\text{init}}$. In particular, we consider different sample sizes $S_{\text{init}} = \left[100, 150, 200, 500 \right]$. The overall training sample size $\left(S_{\text{SDD}}\right)$ is considered to be 1100. The variation of mean relative $L_2$ error with $S_{\text{init}}$ is shown in Fig. \eqref{fig_cs_burger}(a). While the effect of initialization is evident, we note that the slope is relatively flat, indicating the model's mild sensitivity to  $S_{\text{init}}$. Similarly, we also investigate the effect of sample size $S_{\text{SDD}}$ on the performance of the model as shown in Fig. \eqref{fig_cs_burger}(b). We consider $S_{\text{SDD}} = [100,150,200,500,1100]$. For all the cases, we initialize the algorithm from the same initial conditions. As expected, with an increase in the number of training samples, the performance improves as indicated by the reduction in the error. We note that the slope here is relatively more as compared to the previous case study involving $S_{\text{init}}$, indicating the higher sensitivity of the trained model to the number of training samples available in $S_{\text{SDD}}$.


\subsection{1D Wave Advection Equation}\label{sec:AD_eq}
\begin{figure}[ht!]
	\centering
        \includegraphics[width=\textwidth]{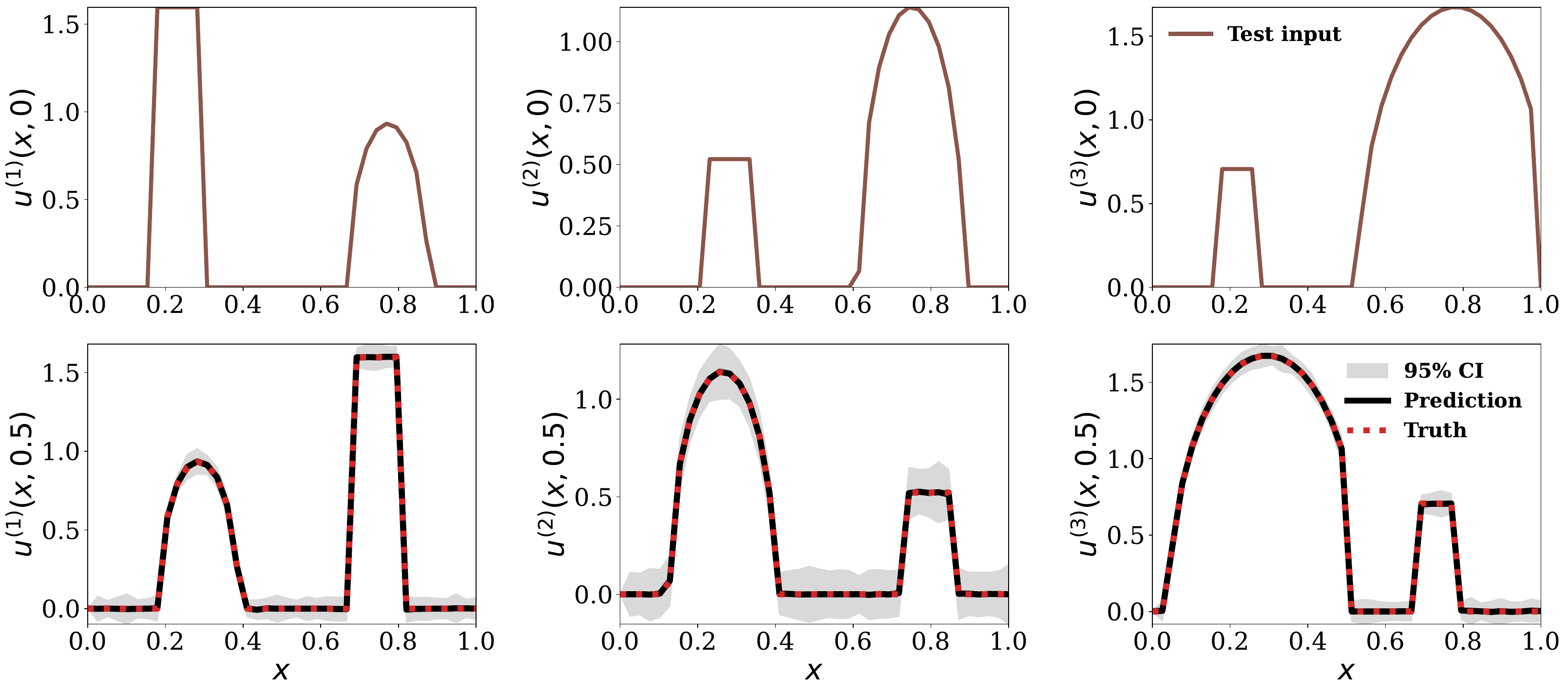}
	\caption{\textbf{1D Wave advection}. Figure illustrates the results obtained from our proposed framework for three of the representative test samples. The first row shows three representative test samples (initial condition) from the test dataset, while the second row shows the corresponding ground truth and the mean prediction with a 95 \% Confidence interval.}
	\label{fig_WA}
\end{figure}

As the second example, we consider the wave advection equation. The equation is frequently used to simulate how waves behave in a variety of mediums, including sound waves in the atmosphere and water waves in the ocean. Periodic boundary conditions are used in this example to mimic the medium's behavior over time. The wave advection equation, when subject to periodic boundary conditions, can be formally formulated as:
\begin{equation}
    \begin{aligned}
    \partial_t u(x,t) + \nu \partial_x u(x,t) &= 0, & & x \in (0,1), t\in (0,1) \\
    u(x - \pi) &= u(x+\pi), & & x \in (0,1).
    \end{aligned}
\end{equation}
Here $\nu $ is positive and greater than zero, $u$ representing the speed of the flow. Here the domain is $(0,1)$ with a periodic boundary condition. The initial condition is chosen as,
\begin{equation}\label{wave_init}
        u(x,0) = h\mathbb{I}_{ \left\{c- \frac{\omega}{2}, c+\frac{\omega}{2} \right\} } + \sqrt{ \max(h^2 - (a(x-c))^2,0) }.
\end{equation}
Here $\omega$ represents the width, $h$ represents the height of the square wave, and $\mathbb{I}_{ \left\{c- \frac{\omega}{2}, c+\frac{\omega}{2} \right\} }$ is an indicator function which takes the value of 1 when $x \in \left\{c- \frac{\omega}{2}, c+\frac{\omega}{2} \right\}$ otherwise 0. The wave is centred around at $x = c$ and the values of \{c,$\omega$,$h$\} are selected from $[0.3,0.7]\times [0.3,0.6] \times [1,2]$. We have considered $\nu =1 $ and a spatial resolution of 40. We aim to learn the mapping from the initial condition to the final condition at $ t = 0.5$, $ u(x,t=0) \mapsto u(x,t=0.5)$. For this example, the proposed framework utilizes 1000 instances of training inputs and is tested on 200 instances of test input. The model was initialized following the same procedure as the previous example. For optimization of the proposed operator using SDD, we utilized a batch size of 20, a learning rate of 0.1, and performed a total of 400 epochs. Additional details regarding the hyperparameter settings are provided in Table \ref{table_accuracy}.
Fig. \ref{fig_WA} illustrates the predictions obtained from the proposed framework. It displays predictions for three different initial conditions, with the ground truth obtained from a numerical solver and the mean predictions from our framework. Fig. \ref{fig_WA} indicates that our proposed framework can produce accurate predictions along with an uncertainty estimate. The uncertainty estimate is shown in the form of a $95 \%$ confidence interval encompassing the mean predictions. From Table \ref{table_accuracy} we observe that our proposed framework gives excellent predictions and outperforms vanilla WNO and GP with vanilla kernel. Again, the effectiveness of SDD is evident from the fact that the proposed framework with SDD outperforms the one without SDD.

\begin{figure}[ht!] 
	\centering 
        \includegraphics[scale = 0.40]{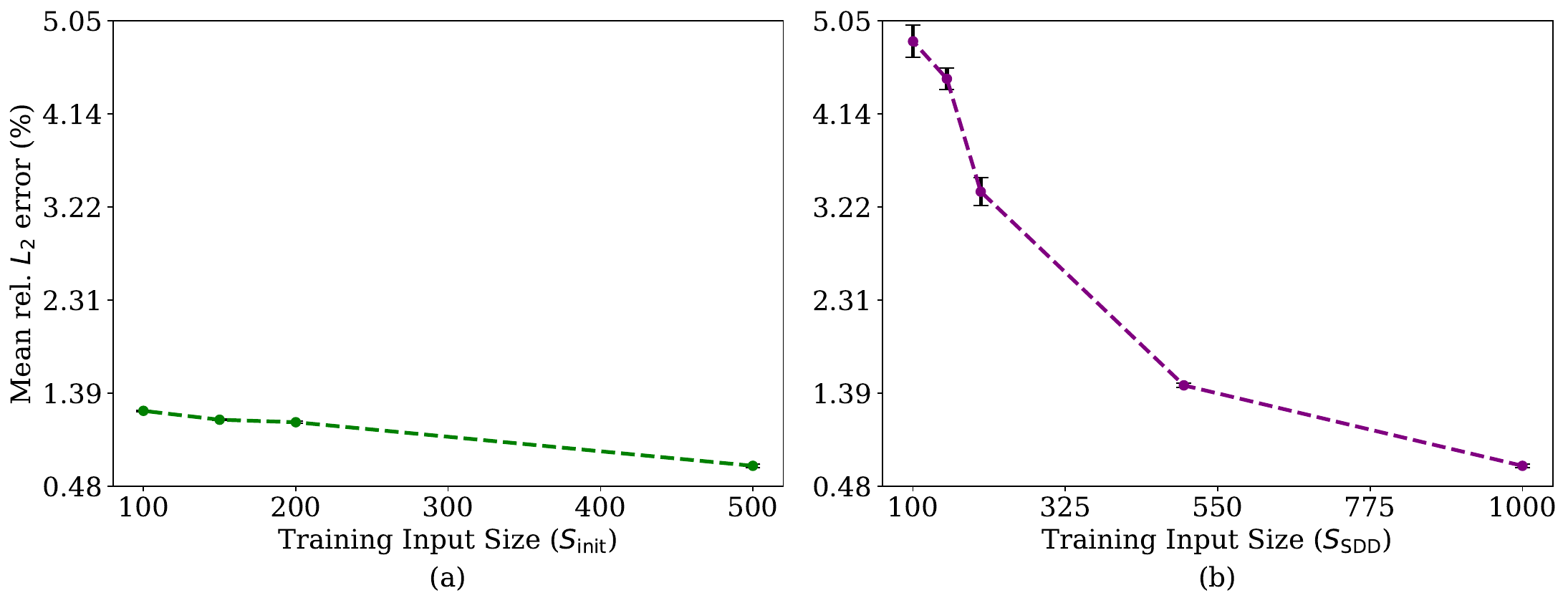} 
	\caption{\textbf{1D Wave advection}. Figure (a) shows the effect of initialization on the performance of the proposed model. We vary the number of samples $S_{\text{init}}$ used for initializing the model while keeping overall training samples $S_{\text{SDD}} = 1000$.  Figure (b) illustrates the effect of the number of training samples $S_{\text{SDD}}$ on the performance of the proposed approach. For all the cases, the model is initialized with the same initial guess.} 
	\label{fig_cs_wa} 
\end{figure}

As with the previous case study for Burger's equation, we analyze the effect of hyperparameters on the performance of the proposed framework. Specifically, we examine the impact of initialization and the number of training samples on the performance of the proposed method. Initialization of the proposed model is done by first training it on a subset of samples ($S_{\text{init}}$). We investigated the effect of initialization by varying $S_{\text{init}} = [100, 150, 200, 500]$. The total number of training samples $(S_{\text{SDD}})$ is considered to be 1000. Fig. \ref{fig_cs_wa}(a) shows the variation of the mean relative $L_2$ error with  $S_{\text{init}}$. Further, to investigate the effect of $S_{\text{SDD}}$ on the performance of the model we consider $S_{\text{SDD}} = [100,150,200,500,100]$. We observe that the performance improves with an increase in the number of the training sample size. We also observe in Fig. \ref{fig_cs_wa} that the sensitivity of the trained model to $S_{\text{SDD}}$ is higher compared to $S_{\text{init}}$.

\begin{table}[ht]
\centering
\caption{Relative $L^2$ Error between the ground truth and the predicted results for the test set.}
\label{table_accuracy}
\begin{threeparttable}
\begin{tabular}{lcccc}
\hline
\multirow{2}{*}{} & \multicolumn{4}{c}{\textbf{Frameworks}} \\ \cline{2-5}
\textbf{PDEs} & \textbf{GPO (with SDD)} & \textbf{GPO (without SDD)} & \textbf{WNO} & \textbf{vanilla GP} \\
\hline
\textbf{Burger}$^{(\ref{sec:bur_eq})}$ & $\mathbf{\approx 2.89 \pm 0.11 \%}$ &$ \approx$  4.62 $\pm$ 0.42 \% & $\approx$ 4.09 $\pm$ 0.21 \% & $\approx$ 4.91 $\pm$ 0.23 \% \\
\textbf{Wave Advection}$^{(\ref{sec:AD_eq})}$ & $\mathbf{\approx  0.63 \pm 0.08 \%}$ & $\approx$  0.98 $\pm$ 0.21 \% &$\approx$ 1.01 $\pm$ 0.11 \% & $\approx$ 1.24 $\pm$ 0.22 \%   \\
\textbf{Darcy}$^{(\ref{sec:Darcy_eq})}$ & $\approx$  3.98 $\pm$ 0.12 \% & $\approx$  5.87 $\pm$ 0.18 \% &$\mathbf{\approx 2.82 \pm 0.44 \%}$ & $\approx$ 6.27 $\pm$ 0.12 \%  \\
\textbf{Darcy (with notch)}$^{(\ref{sec:darcy_td})}$ & $\mathbf{\approx  2.18 \pm 0.19 \%}$ & $\approx$  4.68 $\pm$ 0.29 \% &$\approx$ 2.21 $\pm$ 0.10 \% & $\approx$ 5.22 $\pm$ 0.34 \%\\
\textbf{Navier Stokes}$^{(\ref{sec:NS_eq})}$ & $\mathbf{\approx 2.21 \pm 0.44 \%}$ & $\approx$  7.24 $\pm$ 0.32 \% &$\approx$ 12.23 $\pm$ 0.62 \% &$\approx$ 9.12 $\pm$ 0.43 \% \\
\hline
\end{tabular}
\end{threeparttable}
\end{table}

\subsection{2D Darcy's flow equation}\label{sec:Darcy_eq}
\begin{figure}[ht!]
	\centering
        \includegraphics[width=\textwidth]{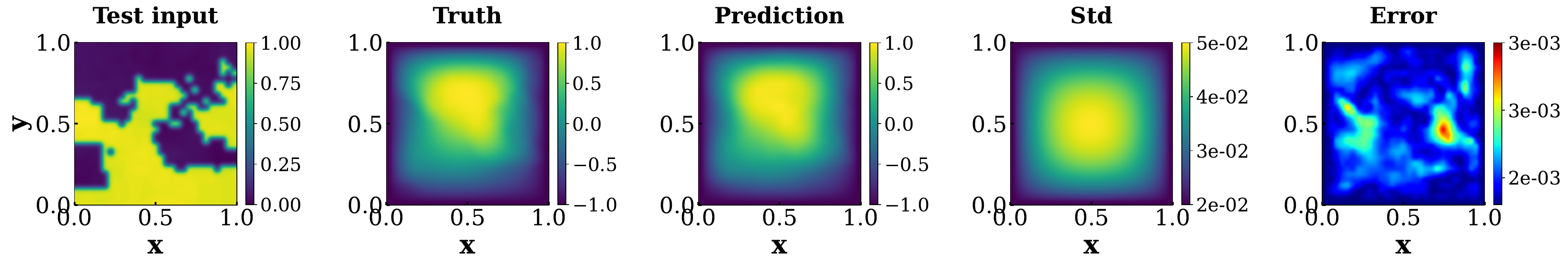}
	\caption{\textbf{2D Darcy flow in a rectangular domain}. Figure shows the results obtained from our proposed framework for one of the representative test inputs. Starting from the leftmost, the plot shows one of the representative test samples of the permeability field, corresponding to the ground truth pressure field, the mean predicted pressure field, the standard deviation of the prediction, and the absolute error.}
	\label{fig_darcy2d}
\end{figure}

\begin{figure}[ht!]
	\centering
        \includegraphics[width=\textwidth]{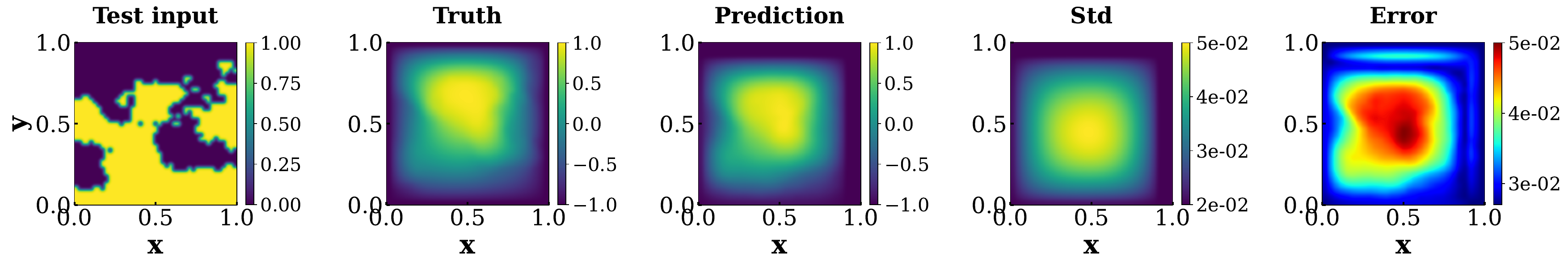}
	\caption{Figure shows zero-shot super-resolution for one of the representative test samples. Starting from the leftmost plot it shows the test input of resolution $53\times53$, the ground truth, the mean and standard deviation of the prediction, and the corresponding absolute error. The model is trained on a spatial resolution of $29\times29$ while it can make predictions on a test input of higher resolution.}
	\label{fig_darcy_super}
\end{figure}

The Darcy flow equation is often used to simulate the movement of fluids through a permeable medium. In this numerical example, we consider the 2D Darcy flow in a rectangular domain to illustrate the efficacy of the proposed framework. The governing equation for the same is represented as:
\begin{equation}
\begin{aligned}
-\nabla \cdot (a(x,y)\nabla u(x,y)) &= f(x,y), & & x \in [0,1], \quad y \in [0,1] \\
u_{\text{BC}}(x,y) &= 0, && x,y \in \partial\Omega
\end{aligned}
\end{equation}
where the Dirichlet boundary condition is set to $u_{\text{BC}}(x,y)=0$. $a(x,y)$ and $u(x,y)$, respectively represent the permeability and pressure fields of the porous medium. The source term is denoted by $f(x,y)$ and its value is set to 1. The aim is to learn the mapping from the permeability field to the solution i.e. $a(x,y) \mapsto u(x,y)$. Similar to the previous example, the proposed framework uses 1000 instances of training inputs and is tested on 200 samples from the test dataset. The parameters of the WNO-embedded kernel were initialized using the same procedure discussed before. In this example, we use a batch size of 16, a learning rate of 0.01, and a total of 300 epochs. Additional details regarding hyperparameters are summarised in Table \ref{table_experimental_settings}.
Fig. \ref{fig_darcy2d} illustrates the results obtained from our proposed framework. It shows predictions for one of the representative permeability fields from the test dataset, along with the ground truth obtained from a numerical solver and the mean predicted pressure field from our framework. As our framework provides a probabilistic solution thus we also obtain the standard deviation of predicted pressure fields. For this example, the proposed approach yields the second-best result after vanilla WNO. having said that, the results obtained using the proposed approach are also accurate as indicated by the low error in Table \ref{table_accuracy}. 
We also investigate the `zero-shot super-resolution' capability of the proposed framework, which refers to its ability to make prediction on a test input of higher spatial resolution even though the model is trained on inputs of lower spatial resolution. The corresponding results are shown in
Fig. \ref{fig_darcy_super}. We observe that even on higher resolution, the results obtained are quite accurate. Additionally, similar to Burger's equation, the predictive uncertainty increases for this case because of the availability of information at new grid points during testing. Overall,  Fig. \ref{fig_darcy_super} establishes the `zero-shot super-resolution' capabilities of the proposed approach.

\subsection{2D Darcy's flow equation with a notch in a triangular domain}\label{sec:darcy_td}
\begin{figure}[ht!]
	\centering
        \includegraphics[width=\textwidth]{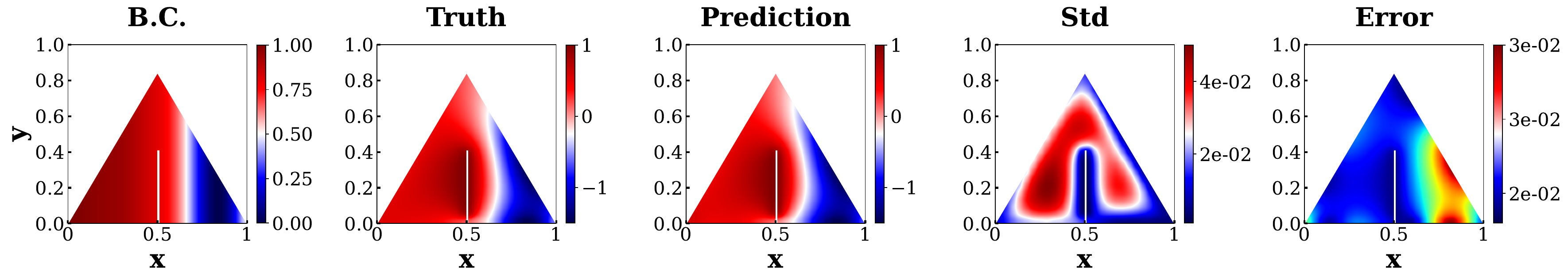}
	\caption{\textbf{2D Darcy flow with a notch in the triangular domain}. Figure shows the results of the flow obtained from our proposed framework for one of the representative test inputs. Starting from the leftmost, the plot shows one of the representative test samples of the boundary condition, corresponding to the ground truth pressure field, the mean predicted pressure field obtained from our proposed framework, the standard deviation of the prediction, and the absolute error.}
	\label{fig_darcy2d_notch}
\end{figure}

As a follow-up to the previous Darcy's flow example, we consider Darcy's flow equation with a notch in the triangular domain. The challenge associated with this problem stems from the presence of the notch and the triangular domain.  
We assume that the boundary condition is sampled from a Gaussian Process,
\begin{equation}
    \begin{aligned}
    u(x,y)\vert_{\partial \Omega} \sim \mathcal{GP}(0,k(x,y,x',y')), && x,y \in [0,1]\\
    k(x,y,x',y') = \exp\left( -\left( \frac{(x - x')^2}{2l_x^2} + \frac{(y - y')^2}{2l_y^2} \right) \right), & & l_x = l_y =0.2, 
    \end{aligned}
\end{equation}
and the objective here is to learn the mapping from the boundary conditions to the pressure fields i.e. $ u(x,y)\vert_{\partial \Omega} \mapsto u(x,y)$. 
The forcing function $f(x,y)$ is set to be a constant and equal to -1 and the permeability field $a(x,y)$ is set to a constant equal to 0.1.
We utilize 1000 instances of training inputs and the proposed framework is tested on 200 instances of test inputs. The WNO-embedded kernel is initialized using the same procedure discussed before. For training the model using SDD, we used a batch size of 8, a learning rate of 0.001, and performed a total of 200 epochs. Table \ref{table_accuracy} provides a summary of the additional details regarding hyperparameter settings.  
Fig. \ref{fig_darcy2d_notch} shows the results obtained from our proposed framework. It displays predictions for one of the representative test inputs from the test dataset. The mean prediction obtained using the proposed approach matches well with the ground truth data generated using numerical simulation. 
As an added advantage of the probabilistic framework, we also get an estimate of the predictive uncertainty that can be utilized for decision-making and adaptive sampling. 
From Table \ref{table_accuracy}, we observe that the proposed approach yields the best result from WNO. The advantage of using SDD is also apparent from the the fact that proposed approach with SDD yields significantly reduced error as opposed to the one without SDD. This gain in accuracy is primarily attributed to the fact that SDD allows the model to scale and utilize more data during training. Vanilla training, on the other hand, suffers from the curse of dimensionality and does not scale with an increase in training data.

\subsection{2D Navier Stokes }\label{sec:NS_eq}
\begin{figure}[ht!]
	\centering
        \includegraphics[width=\textwidth]{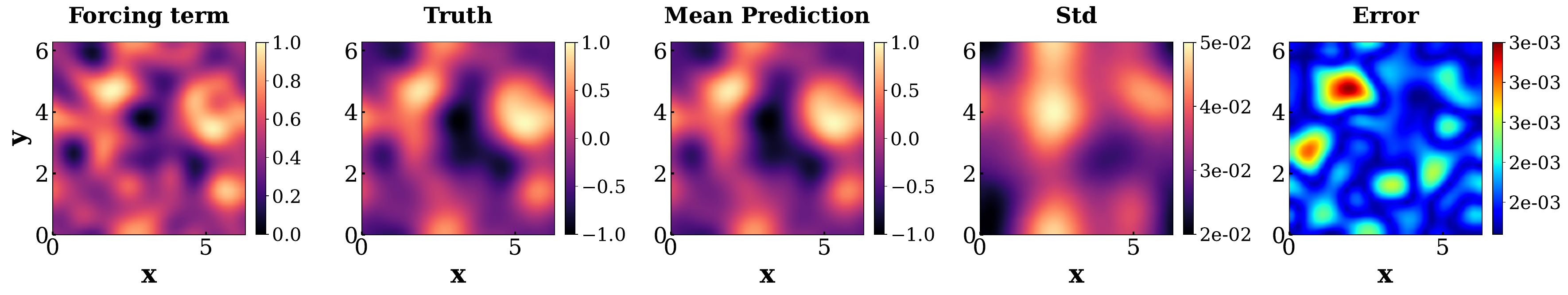}
	\caption{\textbf{2D Navier stokes equation}. Figure shows the prediction obtained from the proposed framework for one of the representative test samples. Starting from the leftmost plot it shows the test input, the ground truth, the mean, and the standard deviation of the prediction, and the corresponding absolute error.}
	\label{fig_NS}
\end{figure}

\begin{figure}[ht!]
	\centering
        \includegraphics[width=\textwidth]{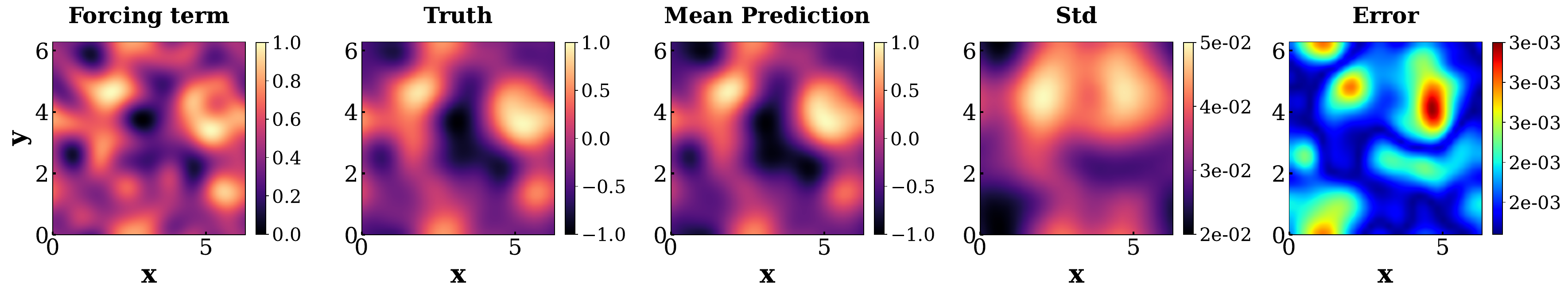}
	\caption{Figure shows zero-shot super-resolution for one of the representative test samples. Starting from the leftmost plot it shows the test input of resolution $64\times64$, the ground truth, the mean and standard deviation of the prediction, and the corresponding absolute error. The model is trained on a spatial resolution of $32\times32$ while it can make predictions on a test input of higher resolution.}
	\label{fig_NS_super}
\end{figure}

As the last example, we consider the Navier-Stokes equation. This equation is essential for understanding physical phenomena, ranging from weather patterns to airflow over airfoils, and ocean currents. For this problem, we consider the vorticity-stream ($\omega$,$\xi$) formulation of the incompressible Navier-Stokes equations,
\begin{equation}
\begin{aligned}
    &\frac{\partial \omega}{\partial t} + (c \cdot \nabla)\omega - \nu \Delta \omega = f', \quad \omega = -\Delta \xi, \\
    &\int_D \xi = 0, \quad c = \left( \frac{\partial \xi}{\partial x_2}, -\frac{\partial \xi}{\partial x_1} \right),
\end{aligned}
\end{equation}
where the periodic domain considered is over $[0,2\pi]^2$. {Here, the forcing term $f'$ is considered to be centered Gaussian with covariance $ \mathbf{C} = (-\Delta + 9\mathbf{I})^{-4}$ and mean function is taken to be zero, where $-\Delta$ denotes the Laplacian operator. The initial condition $\omega(\cdot,0)$ is considered fixed and generated from the same distribution. The viscosity $\nu$ is also fixed to a value of 0.025 and the final time $T = 10$.} Here, we are interested in learning the mapping between the forcing term $f'$ to the vorticity field $\omega(\cdot,T)$. For this example, we have used 3000 instances of training samples, and the proposed framework is tested on 200 test samples. For training of the proposed framework, we utilized a batch size of 16, a learning rate of 0.1, and performed a total of 200 epochs. Further details regarding the hyperparameter settings are provided in the Table \ref{table_accuracy}.
Fig. \ref{fig_NS} illustrates the results obtained using our proposed framework. It shows the prediction for one of the representative test inputs from the test dataset along with the ground truth obtained using a numerical solver and the mean predicted vorticity field accompanied by the standard deviation of the mean prediction. A good match between the two is observed. From Table \ref{table_accuracy}, we observe that our framework gives excellent predictions and outperforms vanilla WNO and GP with vanilla kernel with significant improvement in terms of predictive accuracy. We also investigate the capability of the proposed approach in handling different resolutions during prediction. The results obtained are shown in Fig. \ref{fig_NS_super}. We observe that the proposed framework yields excellent results with the contours matching almost exactly with the ground truth data generated using numerical simulation. Additionally, similar to Burger's equation and Darcy's flow problem, the predictive uncertainty increases for the finer resolution.

\section{Discussions and Conclusion}\label{Conclusion}
In this work, we proposed a novel probabilistic operator learning algorithm, referred to as Gaussian Process Operator (GPO) for computational mechanics. Given the inherent probabilistic nature of Gaussian Processes (GPs), there has been interest among researchers to extend GPs to learn operators including work by Magnani \textit{et al.} \cite{magnani2024linearizationturnsneuraloperators} that proved that linearizing neural operator results in function-valued GP. 
our recent work on Neural Operator induced Gaussian Process (NOGaP) \cite{kumar2024neuraloperatorinducedgaussian} also is an attempt in this direction where a neural Operator was used as a mean function within GP. However, NOGaP is resolution dependent, a key bottleneck as opposed to existing deterministic operator learning algorithms. Additionally, GP has poor scalability with an increase in training samples.  Overcoming these obstacles is vital for extending the applicability of GPs to a broader range of practical problems. The proposed GPO tackles these challenges by offering reliable uncertainty quantification while maintaining computational efficiency, even with large datasets. 
The key advantage of our framework lies in its ability to provide uncertainty-aware predictions, a critical feature in domains such as climate modeling, medical science, and material science, where understanding the uncertainty associated with predictions is as important as the predictions themselves. By utilizing the proposed WNO-embedded kernel design, GPO achieves input-resolution independence.  
Additionally, the proposed approach utilizes stochastic dual descent (SDD) that addresses the bottleneck associated with the scalability of GP and allows GPO to handle large datasets. We 
also provide a practical approach to initialize the SDD algorithm.

In summary, the key findings from this work include:
\begin{itemize}
 \item The incorporation of SDD optimization with GPs provides a scalable and efficient approach for handling large datasets, overcoming a key drawback of GPs.
\item The utilization of WNO embedded kernel design considerably improves the prediction accuracy, making this approach particularly useful for solving parametric PDEs.
\item The framework's probabilistic foundation ensures reliable quantification of predictive uncertainty, which is crucial for several real-world applications.
\end{itemize}

Despite the advantages highlighted above, there still exists certain bottlenecks. For example, the scalability of GPO needs further improvement. One possibility in this direction is to include the inducing points through the sparse GP formulation. An orthogonal research direction is also to develop a physics-informed training algorithm for the proposed GPO. This can potentially address the fact that in computational mechanics, we often deal with scenarios with small datasets. In the future, we will work towards addressing some of these challenges.


\section*{Acknowledgements}
SK acknowledges the support received from the Ministry of Education (MoE) in the form of Research Fellowship. RN acknowledges the financial support received from the Science and Engineering Research Board (SERB), India via grant no. SRG/2022/001410. SC acknowledges the financial support received from Anusandhan National Research Foundation (ANRF) via grant no. CRG/2023/007667. SC and RN acknowledges the financial support received from the Ministry of Port and Shipping via letter no. ST-14011/74/MT (356529). 

\section*{Code availability}
On acceptance, all the source codes to reproduce the results in this study will be made available to the public on GitHub by the corresponding author.


\newpage
\appendix
\section{Wavelet Neural Operator}\label{appendix_WNO}
Neural operators represent a class of deep learning algorithms designed to map functions between infinite-dimensional spaces, unlike traditional artificial neural networks (ANNs) that operate within finite-dimensional vector spaces. The wavelet neural operator is one such example of a neural operator that leverages the wavelet transform's ability to capture both spatial and frequency characteristics, enabling it to learn features effectively in the wavelet domain.

The operations within the wavelet block can be mathematically described as:
\begin{equation}
v_{j+1}(x) = \varphi \left( \left(\mathcal{K}(a;\phi) * v_{j}(x)\right) + b v_{j}(x) \right), \quad j = 1, \ldots, J
\end{equation}
In this expression, $\varphi$ is the non-linear activation function, $\mathcal{K}(\bm{a};\phi)$ denotes the kernel integral operator parameterized by $\phi$, and $v_j(x)$ represents the transformed function from the previous layer. The term $b$ acts as a linear transformation that adjusts the output before applying the non-linearity. The kernel integral operator $\mathcal{K}(a;\phi)$ is defined using a Hammerstein-type integral equation, which is expressed as:
\begin{equation}
\left(\mathcal{K}(a;\phi) v_j\right)(x) = \int_{\Omega} \kappa(x-y;\phi)v_j(y) \, dy
\end{equation}
Here, the kernel function $\kappa(x-y;\phi)$ is parameterized and learned in the wavelet domain.
In the WNO framework, the wavelet transform plays a crucial role. Let $g(x)$ be an orthonormal mother wavelet, which is localized in both the time and frequency domains. The wavelet transform of a function $\xi: \Omega \mapsto \mathbb{R}^d$ is denoted by $\mathcal{W}(\xi)$, while $\mathcal{W}^{-1}(\xi_w)$ represents the inverse wavelet transform. These transforms are given by:
\begin{subequations}
\begin{align}
\mathcal{W}(\xi)(x) &= \int_{\Omega} \xi(x) \frac{1}{|s|^{1/2}} g\left(\frac{x-t}{s}\right) dx \\
\mathcal{W}^{-1}\left(\xi_w\right)(x) &= \frac{1}{C_g} \int_0^{\infty} \int_{\Omega} \xi_w(s, t) \frac{1}{|s|^{1/2}} \tilde{g}\left(\frac{x-t}{s}\right) dt \frac{ds}{s^2}
\end{align}
\end{subequations}
In these equations, $s$ and $t$ are the scaling and translation parameters, $\xi_w$ denotes the wavelet coefficients of the function $\xi$, and $\tilde{g}$ is the dual of the mother wavelet $g$. The constant $C_g$ ensures that the wavelet transform is invertible and is defined as:
\begin{equation}
C_g = 2 \pi \int_{\Omega} \frac{\left|g_\omega\right|^2}{|\omega|} d \omega
\end{equation}
where $g_\omega$ is the Fourier transform of $g$.
Finally, by leveraging the convolution theorem \cite{perez2004convolution}, WNO can be expressed as:
\begin{equation}
(\mathcal{K}(\phi) \ast v_j)(x) = \mathcal{W}^{-1} \left( R_\phi \cdot \mathcal{W}(v_j) \right)(x)
\end{equation}
Here, $R_\phi$ represents the wavelet-transformed kernel, parameterized by $\phi$. For more details, interested readers can refer to \cite{tripura2023wavelet}

\end{document}